\documentclass[letterpaper, 10 pt, conference]{ieeeconf}  

\IEEEoverridecommandlockouts                              

\usepackage{cite}
\usepackage{amsmath,amssymb,amsfonts}
\usepackage{algorithmic}
\usepackage{algorithm}
\usepackage{graphicx}
\usepackage{textcomp}
\usepackage{xcolor}
\usepackage[acronyms]{glossaries}
\usepackage{mathtools}
\usepackage[colorinlistoftodos,prependcaption]{todonotes}
\usepackage{regexpatch}
\usepackage{svg}
\usepackage{enumerate}
\makeatletter
\xpatchcmd{\@todo}{\setkeys{todonotes}{#1}}{\setkeys{todonotes}{inline,#1}}{}{}
\makeatother
\usepackage{tikz-network}
\usepackage{booktabs}
\usepackage{flushend} 
\usepackage{caption}
\usepackage{subcaption}
\usepackage{comment}
\usepackage{jabbrv}
\usepackage{multirow}
\usepackage{hyperref}

\def\BibTeX{{\rm B\kern-.05em{\sc i\kern-.025em b}\kern-.08em
    T\kern-.1667em\lower.7ex\hbox{E}\kern-.125emX}}

\def\ie{\textit{i.e.},}
\def\eg{\textit{e.g.},}
\def\sota{state-of-the-art}

\newcommand{\qt}[1]{``#1''}

\def\figvspace{\vspace{-1.5em}}
\def\tabcapspace{\vspace{-1em}}

\newacronym{llm}{LLM}{Large Language Model}
\newacronym{vlm}{VLM}{Visual-Language Model}
\newacronym{vit}{ViT}{Visual Transformer}
\newacronym{lm}{LM}{Language Model}
\newacronym{nn}{NN}{Neural Network}
\newacronym{cnn}{CNN}{Convolutional Neural Network}
\newacronym{lsm}{LSM}{Latent Semantic map}
\newacronym{mcl}{MCL}{Monte-Carlo Localization}
\newacronym{ndt}{NDT}{Normal Distributions Transform}
\newacronym{cndt-om}{C-NDT-OM}{Clustered Normal Distributions Transform Occupancy Map}
\newacronym{ndt-mcl}{NDT-MCL}{Normal Distributions Transform Monte-Carlo Localization}
\newacronym[plural=NDT-OMs,firstplural=Normal Distributions Occupancy Maps (NDT-OM)]{ndt-om}{NDT-OM}{Normal Distributions Occupancy Map}
\newacronym{ndt-se}{NDT-SE}{Semantically Enhanced Normal Distributions Transform}
\newacronym{d2d}{D2D}{Distribution-to-Distribution}
\newacronym{p2d}{P2D}{Point-to-Distribution}
\newacronym{ins}{INS}{Inertial Navigation System}
\newacronym{ned}{NED}{North-East-Down}
\newacronym{utm}{UTM}{Universal Transverse Mercator}
\newacronym{rmse}{RMSE}{Root Mean Squared Error}
\newacronym{ate}{ATE}{Absolute Trajectory Error}
\newacronym{rpe}{RPE}{Relative Pose Error}
\newacronym{hmm}{HMM}{Hidden Markov Model}
\newacronym{av}{AV}{autonomous vehicle}
\newacronym{ros}{ROS}{Robot Operating System}
\newacronym{gps}{GPS}{Global Positioning System}
\newacronym{gnss}{GNSS}{Global Navigation Satellite System}
\newacronym{sd-ndt-om}{SD-NDT-OM}{Semantic-Dynamic Normal Distributions Occupancy Map}
\newacronym{slam}{SLAM}{Simultaneous Localization and Mapping}
\newacronym{cow}{CoW}{Clip on Wheels}
\newacronym{iou}{IoU}{Intersection over Union}

\title{\LARGE \bf
Do Visual-Language Grid Maps Capture Latent Semantics? 
}

\author{Matti Pekkanen, Tsvetomila Mihaylova, Francesco Verdoja, and Ville Kyrki 
\thanks{This work was supported by Business Finland (decision 9249/31/2021), the Research Council of Finland (decision 354909), Wallenberg AI, Autonomous Systems and Software Program, WASP and Saab AB. We gratefully acknowledge the support of NVIDIA Corporation with the donation of the Titan Xp GPUs used for this research.}
\thanks{M. Pekkanen, T. Mihaylova, F. Verdoja and V. Kyrki are with School of Electrical Engineering, Aalto University, Espoo, Finland. {\tt\small \{firstname.lastname\}@aalto.fi}}%
}

\begin{document}

\maketitle
\thispagestyle{empty}
\pagestyle{empty}

\bstctlcite{IEEEexample:BSTcontrol}

\begin{abstract}
Visual-language models (VLMs) have recently been introduced in robotic mapping using the latent representations, \ie{} embeddings, of the VLMs to represent semantics in the map. They allow moving from a limited set of human-created labels toward open-vocabulary scene understanding, which is very useful for robots when operating in complex real-world environments and interacting with humans. While there is anecdotal evidence that maps built this way support downstream tasks, such as navigation, rigorous analysis of the quality of the maps using these embeddings is missing. 
In this paper, we propose a way to analyze the quality of maps created using VLMs. We investigate two critical properties of map quality: queryability and distinctness. The evaluation of queryability addresses the ability to retrieve information from the embeddings. We investigate intra-map distinctness to study the ability of the embeddings to represent abstract semantic classes and inter-map distinctness to evaluate the generalization properties of the representation. 
We propose metrics to evaluate these properties and evaluate two state-of-the-art mapping methods, VLMaps and OpenScene, using two encoders, LSeg and OpenSeg, using real-world data from the Matterport3D data set. Our findings show that while 3D features improve queryability, they are not scale invariant, whereas image-based embeddings generalize to multiple map resolutions. This allows the image-based methods to maintain smaller map sizes, which can be crucial for using these methods in real-world deployments. Furthermore, we show that the choice of the encoder has an effect on the results. The results imply that properly thresholding open-vocabulary queries is an open problem.
\end{abstract}
\section{Introduction}
\label{sec:intro}

Robots must understand the geometry and semantics of the environment to accomplish complex tasks. The semantic information stored in the map must be able to be \textit{queried}. While complex queries have been long possible in semantic maps by inference using object ontologies \cite{tenorth_knowrob_2009}, semantic labels support only binary comparison as they are enumerations with no notion of distance. This imposes challenges in real-world settings when a robot operates in an environment that the categories do not fully describe \cite{bendale_towards_2016,geng_recent_2021}.

\begin{figure}[t]
    \centering
    \includegraphics[width=0.475\textwidth]{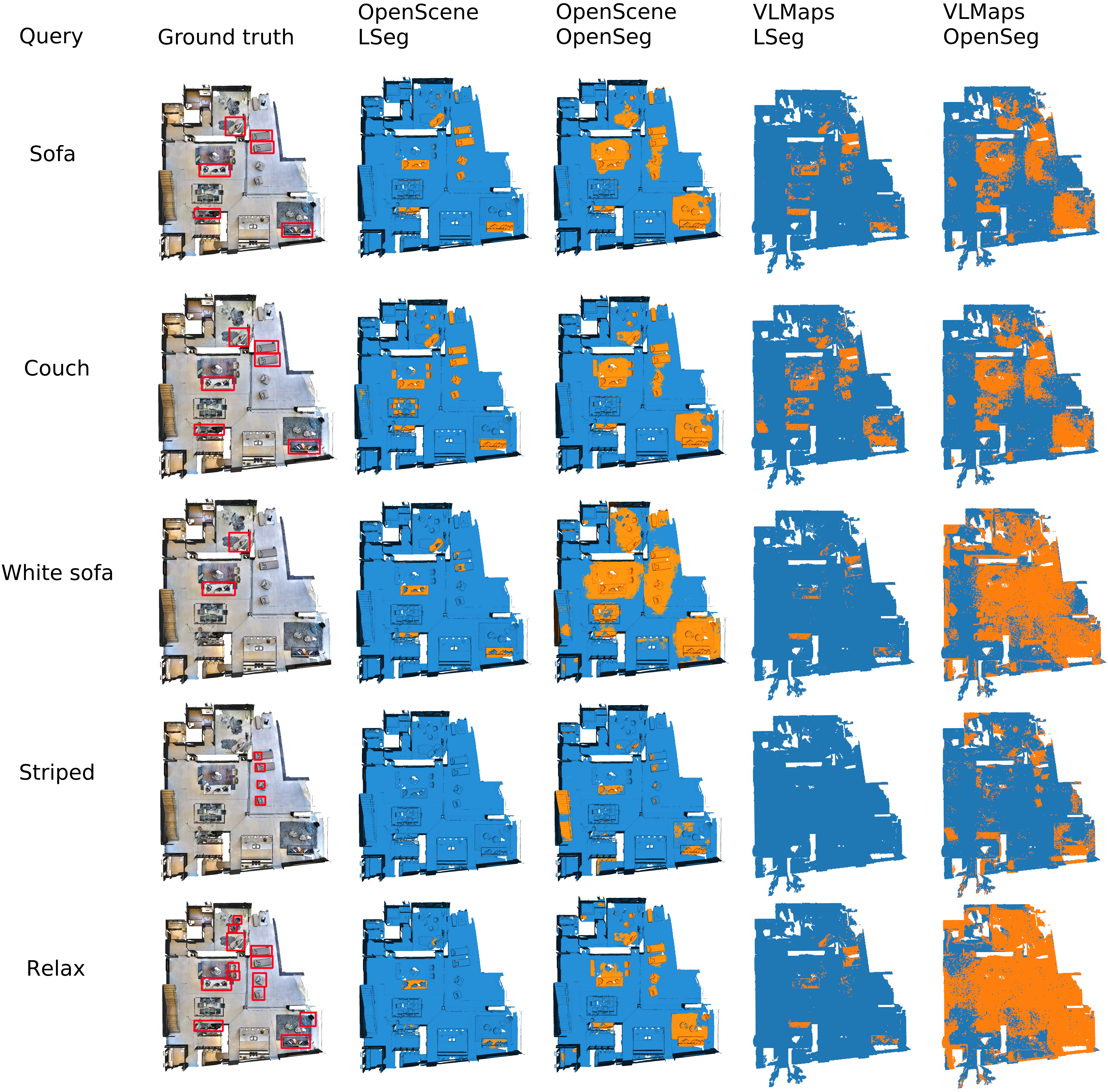}
    \caption{ 
    Maps built from visual-language models have the ability to represent complex semantics, encompassing both the class of objects and their properties. We propose a way to evaluate the quality of different \sota{} methods according to the distinctness of their semantic concepts and their ability to be queried.
    }
    \label{fig:header-img}
    \figvspace{}
\end{figure}

Lately, grid map methods have been proposed that use \gls{vlm} embeddings as semantics, which we will collectively refer to as \textit{latent semantic maps}. Their promise is that queries can be evaluated by directly comparing the query embedding to the embeddings of the map with a similarity metric, yielding virtually unrestricted query vocabulary. Several successful algorithms have been built on \glspl{vlm} that can perform mobile robot tasks, such as navigation \cite{huang_visual_2023, gadre_cows_2023, shah_lm-nav_2022}. However, the analysis from the robotic mapping perspective is left anecdotal.

This work aims to bridge this gap by analyzing the performance of two state-of-the-art grid mapping methods, VLMaps \cite{huang_visual_2023} and OpenScene \cite{peng_openscene_2023}. We propose two metrics for this analysis: queryability and distinctness of semantic concepts. Queryability measures the ability to find matches for queries from the representation, while distinctness measures the ability to distinguish different abstract object classes. Furthermore, we explore the practical considerations of the methods by comparing changes in queryability with varying resolutions, as the map size is a function of the resolution of the map.

The main contributions of this paper are:
\begin{enumerate}[i)]
    \item We define queryability and distinctness of representation as measures of mapping quality for latent semantic maps and propose metrics for quantitatively evaluating the distinctness;
    \item We provide a comprehensive analysis of two state-of-the-art methods using the proposed metrics;
    \item We evaluate the behavior of the maps with different resolutions;
    \item We make our evaluation software publicly available\footnote{\href{https://aalto-intelligent-robotics.github.io/latent_semantics/}{https://aalto-intelligent-robotics.github.io/latent\_semantics/}}, enabling evaluation of the progress in this new field of research.
\end{enumerate}

\section{Related work}
\label{sec:related}

\subsection{Grid maps created with visual-language embeddings}
Creating maps using embeddings of \glspl{vlm}, especially CLIP \cite{radford_learning_2021}, has recently gained attention, with the proposal of methods such as VLMaps \cite{huang_visual_2023}, OpenScene \cite{peng_openscene_2023}, NLMaps \cite{chen_open-vocabulary_2022}, Uni-Fusion \cite{yuan_uni-fusion_2024}, ConceptFusion \cite{jatavallabhula23conceptfusion}, ConceptGraphs \cite{gu2024conceptgraphs}, and HOV-SG \cite{werby24hovsg}.

The most common type of map used in robotics is a dense grid map. VLMaps and OpenScene have augmented these maps with open-vocabulary semantics by associating each cell with a single \gls{vlm} embedding. The embeddings are created from RGBD images with a language-driven semantic segmentation encoder, such as LSeg \cite{li_language-driven_2022} or OpenSeg \cite{ghiasi_scaling_2022}, both based on CLIP, and back-projected to the 3D map. Each cell is represented as the mean of the embeddings projected to the cell. In the original work, the 3D map in VLMaps is projected to 2D by averaging the embeddings on the $z$-axis. Additionally, OpenScene learns an encoder that directly produces CLIP embeddings from the 3D point cloud. Combining these two approaches, OpenScene proposes joint 2D-3D \qt{ensemble} features as the final representations in the map. 

Other methods have been proposed, which do not use grid maps or do not produce embeddings on each grid cell. NLMaps creates a feature-based map, where each feature is associated with an \gls{vlm} embedding. Uni-Fusion proposes a general method to create continuous maps from any perceptual data modality, including \gls{vlm} embeddings. ConceptFusion builds a dense point cloud map with an embedding for each point, whereas ConceptGraphs and HOV-SG have dense point cloud maps with one embedding per object. The extension of the evaluation of representations with object-level embeddings is left for future works.

\subsection{Open-vocabulary queries}
Open-vocabulary methods aim to move beyond predicting a constrained, \ie{} closed, set of labels. This means that not only should the open-vocabulary systems be able to capture synonyms and closely related terms but also attributes and related actions of the objects. This has been explored early on by formalizing the relation of labels into conceptual graphs \cite{zhao_open_2017}, similar to what is used in semantic mapping, predicting attributes of objects \cite{bravo_open-vocabulary_2023} to the prediction of affordances and activities related to objects \cite{peng_openscene_2023}.

There are closely related fields to open-vocabulary detection and segmentation, especially open-set and zero-shot learning. In open-set learning, the task is to classify the known classes seen in training and to classify unseen classes as unseen \cite{geng_recent_2021}. Zero-shot learning aims to be able to classify classes that did not appear in the training data \cite{bansal_zero-shot_2018}. This, however, does not imply that open vocabulary is in use. Additionally, many open-vocabulary methods \cite{zareian_open-vocabulary_2021, gu_open-vocabulary_2022, feng_promptdet_2022}, even when trained with open-vocabulary methods, are evaluated in a closed-vocabulary but zero-shot setting. While this fails to capture the full potential of the open vocabulary, the results are comparable to semantic segmentation methods, where currently, fully supervised methods outperform the visual-language methods \cite{peng_openscene_2023}.

In contrast, in this work, we evaluate queryability such that the evaluation is not restricted to semantic segmentation of the map, which imposes that the map must be partitioned. Instead, each query is evaluated separately, as it would be in real applications.

\subsection{Analysis of quality of maps}
Most existing works assessing the quality of robotic maps concentrate on evaluating geometry using methods ranging from simple feature descriptors \cite{wagan_map_2008}, frequency-based methods \cite{kucner_robust_2021}, to probabilistic measures \cite{aravecchia_comparing_2024}. 

Evaluating the geometric and semantic properties jointly is only relevant when the geometry and semantics are jointly estimated, \eg{} in semantic \gls{slam}. Often, the metrics are qualitative, but quantitative metrics such as the localization accuracy or map reconstruction quality compared to ground truth objects are used as well \cite{salas-moreno_slam_2013, mccormac_fusion_2018, runz_maskfusion_2018}. Like many semantic mapping works, \cite{sunderhauf_meaningful_2017, chen_leveraging_2023, narita_panopticfusion_2019}, we focus on the evaluation of the semantic properties separately, using binary classification metrics.

In VLMaps, as with other zero-shot navigation methods \cite{gadre_cows_2023, shah_lm-nav_2022}, the map quality is not directly assessed, but instead, they use the success of the downstream navigation task as the metric, which does not fully evaluate the capabilities of the representation.

Because the metrics used in evaluating latent semantic maps in each prior work are different, comparing the quality of their maps is non-trivial, which is the problem we address in this paper.
\section{Problem statement}
\label{sec:problem}

In this study, we evaluate the quality of latent semantic maps. We only consider methods where the estimation of geometry and semantics are disjoint, so we concentrate on evaluating the semantic quality of the maps, which we consider to be captured by two properties: queryability and distinctness.

Given a map $m_i$, an open-vocabulary query $q$, and a query function $\varphi$, a binary mask $b^i_q = \varphi(m_i, q)$ is acquired. Given the ground truth binary mask $b^i_t$ for the same query, the \emph{queryability} $Q$ with respect to $q$ and a metric $F_q$ is defined as $Q^i_q = F_q(b^i_q, b^i_t)$.

Given a set $\mathcal{S}^i_l$ of semantic representations of a class $l$ in map $m_i$, \emph{distinctness} $D$ between two labels $l$ and $l'$, from maps $m_i$ and $m_{i'}$, respectively, and with respect to a metric $F_d$, is defined as $D^{i,\, l}_{i',\, l'} = F_d(\mathcal{S}^{i}_{l}, \mathcal{S}^{i'}_{l'})$.

\section{Methods}
\label{sec:methods}

\subsection{Evaluating queryability}
\label{sec:voxel-queryability}

Queryability captures the ability to retrieve information from the embeddings, \ie{} to find the parts of the map that answer the query. The hypothesis is that from a map with high queryability, it is possible to find accurate matches to a wide range of queries.

Queryability is evaluated by comparing the query results and ground truth in a binary classification setting. This property cannot be evaluated using standard multi-class classification, as queries do not form a partition of the map. Each query result is a binary mask over the whole map, and the same voxel might be a match of multiple queries. This is evident from the open-vocabulary perspective; \eg{} a sofa might match queries \qt{sofa}, \qt{couch}, \qt{place to sit}, and \qt{soft}.

The method consists of the following steps:
\begin{enumerate}[i)]
    \item A map $m$ is created from each sequence from the data set, forming the set $\mathcal{M}$.
    \item Each map $m \in \mathcal{M}$ is queried with each query $q$ in the set of queries $\mathcal{Q}$. The query result of a single query is a binary segmentation mask of the map, $\hat{\textbf{y}}_q = \{\hat{y}_1, \dots, \hat{y}_N\}$, consisting of $N$ voxel masks $\hat{y} \in \{ true, false \}$. Combined, the query results form the set of voxel-based predictions $\hat{Y}_V = \{\hat{\textbf{y}}_q | q \in \mathcal{Q}\}$.
    \item For each query, the true mask $\textbf{y}_q$ is created from the ground truth map $m_g$ such, that the voxels answering query $q$ are $true$, others $false$. The true masks form a set of true labels $Y_V= \{\textbf{y}_q | q \in \mathcal{Q}\}$.
    \item The binary classification metrics are calculated between $Y_V$ and $\hat{Y}_V$.
\end{enumerate}

We use precision, recall, f1-score, and \gls{iou} as metrics. We acquire the queryability by averaging them over all of the maps and queries with respect to each metric $Q^{\mathcal{M}}_{\mathcal{Q}} = F_q(Y_V, \hat{Y}_V), \: F_q \in \{ p, r, f_1, iou \}$.

With most queries, the regions of interest are relatively small, so true negatives dominate the predictions. Therefore, accuracy cannot be used as a metric, as it incorporates true negative predictions. The selected metrics are not affected by this problem.

\subsection{Evaluating distinctness}

We further subdivide the distinctness of the map into two distinct aspects, intra-map distinctness, and inter-map distinctness, and propose a method for evaluating each.

\subsubsection{Intra-map distinctness} \label{sec:intra-cons}

Intra-map distinctness captures the similarity of embeddings across the voxels within a map, sharing semantic meaning. The hypothesis is that embeddings sharing semantic meaning are clustered together in the latent space, allowing the distinguishing of different concepts. 

The method consists of the following steps:
\begin{enumerate}[i)]
    \item Given the $i$-th map $m_i$, and semantic label $l$, $\mathcal{E}_l$ is the set of embeddings corresponding to voxels in the map $m_i$ where the ground truth semantic label is $l$, and $\mathcal{E}_i$ is the set of all embeddings in the map $m_i$. We form a set of tuples $\mathcal{T} = \{(\mathcal{E}_l, \mathcal{E}_i) | m_i \in \mathcal{M}, l \in \mathcal{L}\}$, where $\mathcal{L}$ is a closed-set semantic label vocabulary. For computational efficiency and to no considerable change in results, in practice, the sets $\mathcal{E}_l$ and $\mathcal{E}_i$ are subsampled. We subsample each label separately to preserve the label distribution. We use a subsampling ratio of $0.1$ in this work.
    \item The average absolute deviation $d_l^i$ is calculated for $\mathcal{E}_l$, and $d_\mu^i$ for $\mathcal{E}_i$, for each tuple $t \in \mathcal{T}$. The average absolute deviation $d_a$ measures the statistical dispersion of a set, and given a set of points $\mathcal{E} = \{e_1, ..., e_n\}$ it is defined as 
\begin{equation}
    d_a = \frac{1}{n} \sum^n_{j=1} | f_d(e_j, \bar{\mathcal{E}})|,
\end{equation}
where $f_d$ is a deviation metric and $\bar{\mathcal{E}}$ a central point of set $\mathcal{E}$. We use the mean of the embeddings as the central point and cosine distance as the deviation metric, which is consistent with the use of cosine distance loss for CLIP.
    \item The intra-map distinctness ratio $c_l^i = \frac{d_l^i}{d_\mu^i}$ is computed for each tuple $t \in \mathcal{T}$, with label $l$ in map $m_i$. This ratio represents the distinctness $D^{i,\, l}_{i,\, \mathcal{L}}$ of the class $l$ compared to the map average.
\end{enumerate}

\subsubsection{Inter-map distinctness}
Inter-map distinctness captures the similarity of voxels with the same semantic label across different maps. The hypothesis is that embeddings within the same label are closer to each other across maps than to those with different labels with respect to a distance metric, \ie{} $D^{i,\, l}_{j,\, l} < D^{i,\, l}_{i,\, l'}$. This would imply that the object classes retain their distinctness across maps, and therefore, the system generalizes better across different environments. 

The method consists of the following steps:
\begin{enumerate}[i)]
    \item Given the set of tuples $\mathcal{T}$, constructed according to Section \ref{sec:intra-cons}.
    \item For all pairs of tuples $\big((\mathcal{E}_{l_1}, \mathcal{E}_{i_1}), $ $(\mathcal{E}_{l_2}, \mathcal{E}_{i_2})\big)$, ${i_1} \neq {i_2}$, parametric Wasserstein 2-distance $d_w$ is calculated between $\mathcal{E}_{l_1}$ and $\mathcal{E}_{l_2}$, representing distinctness $D^{i_1,\, l_1}_{i_2,\, l_2}$. We subsample $\mathcal{E}_{l_1}$ and $\mathcal{E}_{l_2}$ similarly to Section \ref{sec:intra-cons}.
\end{enumerate}

If the embeddings are normalized, they can be thought of as high-dimensional probability distributions. To measure the distance between the distributions of $\mathcal{E}_{l,1}$ and $\mathcal{E}_{l,2}$, we selected the Wasserstein $p$-distance, which is a common metric of the distance between probability distributions. However, it is computationally heavy for large sets of high-dimensional embeddings. For this reason, given an n-dimensional embedding $e \in \mathbb{R}^n$, we approximate the set of embeddings $\mathcal{E}$ with an $n$-dimensional normal distribution $\mathcal{N}(\mu, P)$. This allows us to have a closed-form solution for the Wasserstein 2-distance
\begin{equation}
    \label{eq:wsd_nd}
    d_w(\mathcal{E}_1, \mathcal{E}_2) \approx \lVert \mu_1 - \mu_2 \rVert^2_2 + \text{tr}\big( P_1 + P_2 - 2 (P_2^{\frac{1}{2}} P_1 P_2^{\frac{1}{2}})^{\frac{1}{2}} \big).
\end{equation}
\begin{figure}[t]
\centering
    \includegraphics[width=0.49\textwidth]{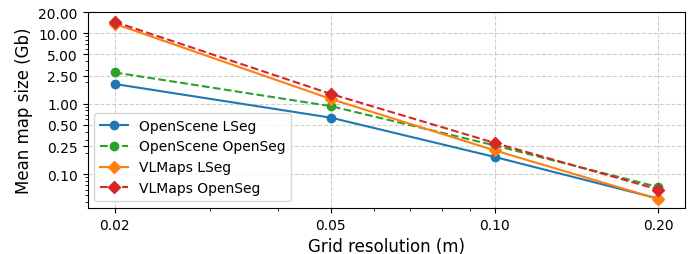}
    \caption{The mean map size in gigabits (Gb) as a function of the map cell size.}
    \label{fig:map-size}
    \figvspace{}
\end{figure}
\begin{table*}[t!]
    \caption{Average results of the queryability tests over 10 maps.}
    \tabcapspace{}
    \begin{center}
        \setlength\tabcolsep{6pt}
        \scalebox{1.0}{
                \begin{tabular}{cccccccccccc}
                    \toprule
                    & & & \multicolumn{4}{c}{Semantic segmentation} & & \multicolumn{4}{c}{VLMaps querying} \\
                    \cmidrule(lr){4-7}\cmidrule(lr){8-12}
                    \textbf{ Method } & \textbf{ Encoder } & & \textbf { F1 } & \textbf{ Precision } & \textbf { Recall } & \textbf { IoU } & & \textbf { F1 } & \textbf{ Precision } & \textbf { Recall } & \textbf { IoU } \\
                    \midrule
                    OpenScene & LSeg    & & \textbf{0.645} & \textbf{0.639} & \textbf{0.651} & \textbf{0.480} & & 0.174	         & 0.096          & 0.897          & 0.095           \\
                    OpenScene & OpenSeg & & 0.603          & 0.598          & 0.609          & 0.435          & & 0.111	         & 0.059          & \textbf{0.975} & 0.059           \\
                    VLMaps    & LSeg    & & 0.541          & 0.529          & 0.553          & 0.375          & & \textbf{0.425} & \textbf{0.297} & 0.757          & \textbf{0.271}  \\
                    VLMaps    & OpenSeg & & 0.495          & 0.526          & 0.469          & 0.333          & & 0.367          & 0.256          & 0.700          & 0.226           \\
                    \bottomrule
                \end{tabular}
        }
        \setlength\tabcolsep{6pt}
        \label{tab:queryability-results}
    \end{center}
    \figvspace{}
\end{table*}
\section{Experiments}
\label{sec:experiments}

The three main questions that we aim to answer with the experiments are:
\begin{enumerate}
    \item How queryable state-of-the-art latent semantic maps are?
    \item How distinct are their visual-language embeddings within and across maps?
    \item How does the grid map resolution affect the queryability?
\end{enumerate}

To answer these questions, we evaluated two state-of-the-art methods, VLMaps \cite{huang_visual_2023} and OpenScene \cite{peng_openscene_2023}, in a series of experiments.

\subsection{Data set}
The data set used in the study is the Matterport3D data set \cite{chang_matterport3d_2017}, which consists of houses captured with an RGBD camera and ground truth semantic segmentation. The data set was used in the original work of VLMaps, in which they selected a sample of 10 sequences from the data set. In this work, we use the same sequences and the same set of 40 labels \cite{chang_matterport3d_2017} forming the set $\mathcal{L}$; their names are also used in the following experiments as the query set $\mathcal{Q}$ to leverage the semantic ground truth provided with the dataset\footnote{\href{https://aalto-intelligent-robotics.github.io/latent_semantics/}{https://aalto-intelligent-robotics.github.io/latent\_semantics/}}.

Since Matterport3D does not provide ground truth instance segmentation, instances used as ground truth were segmented using a region growing algorithm based on the ground truth semantics \cite{adams_seeded_1994}. Each voxel is initialized as a seed cluster; then, region growing steps are performed, where the labels of all neighboring clusters are compared. If the labels are the same, the clusters are joined. Otherwise, they are not. This step is iterated until no more clusters can be joined or a maximum iteration limit is reached.

\subsection{Parametrization of the methods}
Instead of using the 2D mapping method proposed in the original work \cite{huang_visual_2023}, the VLMaps map was created using the new 3D mapping method available at the repository of the original authors \cite{VLMaps_software}. This choice was made to create a fair comparison using state-of-the-art methods. The central idea is the same, except the 3D map avoids the problems when aggregating the embeddings along the $z$-axis. All of the parameters were the default used by the original authors. 

The 3D mapping method was also extended to create semantic ground truth maps from the Matterport3D data. The VLMaps use Habitat simulator \cite{habitat19iccv, szot2021habitat, puig2023habitat3} for camera measurements. Similarly to the RGB images, we create semantic images and backproject them to the 3D environment.

OpenScene maps were created using the parameters proposed in the original paper. The main analysis uses the proposed 2D-3D ensemble features, but both 2D image features (\qt{fusion}) and 3D point cloud features (\qt{distill}) were created for the ablation study. 

Both methods used the same pre-trained LSeg and OpenSeg encoder provided by the original authors. Both encoders are based on the CLIP backbone; LSeg uses the CLIP-ViT-B/32 backbone with $512$-dimensional embeddings, whereas OpenSeg uses the CLIP-ViT-L/14 with $768$-dimensional embeddings.

\subsection{Querying the maps}
VLMaps proposes a method for performing open-vocabulary queries, which we will refer to as \textit{VLMaps querying}. When the map is queried with a query $q$, the embeddings on the map are compared using cosine similarity to two embeddings, embedding of $q$ and embedding of string \qt{other}. The binary classification mask is $true$ for voxels where the most similar query is $q$; otherwise, it is $false$. The binary mask is then post-processed by a sequence of binary closing, gaussian blurring, thresholding, and binary dilation to encompass whole object instances. 

OpenScene does not propose a method to perform open-vocabulary queries. In the original work, the map is evaluated in a semantic segmentation setting, and open-vocabulary queries are evaluated qualitatively. In our experiment, we implement the VLMaps querying method on OpenScene and use the same parametrization for both methods. Additionally, we evaluate both methods in a semantic segmentation setting, where a binary query mask is acquired by comparing the equality of the voxel labels on the map and the query label.

Finally, VLMaps proposed to use prompt engineering by creating a list of queries by setting $q$ and string \qt{other} into a list of phrases such as \qt{A photo of ...} and \qt{there is ... in the scene}. Then, the augmented lists of prompts are embedded, and the embeddings are averaged to acquire two embeddings, one for $q$, one for \qt{other}, for comparison. We evaluate the effect of prompt engineering on queryability on VLMaps.

\subsection{Resolution analysis}
For real-world applications, where the robot hardware capabilities might be limited, the memory footprint of the map can be crucial. Small cell sizes can lead to prohibitively large maps, and the size of a grid map is a function of the resolution of the map, as shown in Figure \ref{fig:map-size}. OpenScene has smaller map sizes due to downsampling the images. To investigate the effects of cell size on the map quality, all of the experiments were performed with a set of resolutions $r \in [ 0.02, 0.05, 0.1, 0.2 ]$ m. The resolutions proposed in the original works are $0.02$m and $0.05$m in OpenScene and VLMaps, respectively. If not explicitly stated otherwise, we show the results in a resolution of $0.02$m for fair comparison.

\section{Results}
\label{sec:results}

\subsection{Queryability}
\label{sec:queryability}

The results of the queryability are presented in Table \ref{tab:queryability-results}, where the F1-score, precision, recall, and \gls{iou} of all of the method and encoder combinations are presented for the semantic segmentation and VLMaps querying. 

\subsubsection{Semantic segmentation}
In semantic segmentation, overall, OpenScene performs better than VLMaps with both encoders, based on the higher F1-scores and \glspl{iou}. Furthermore, LSeg performs better with both methods than OpenSeg. Therefore, these results show that both the mapping method and choice of encoder matter in the map creation process. They also indicate that the 3D structure of the environment, included by OpenScene, contains information that can be leveraged in addition to the purely image-based creation of embeddings. 

\begin{figure}[t]
\centering
    \includegraphics[width=0.49\textwidth]{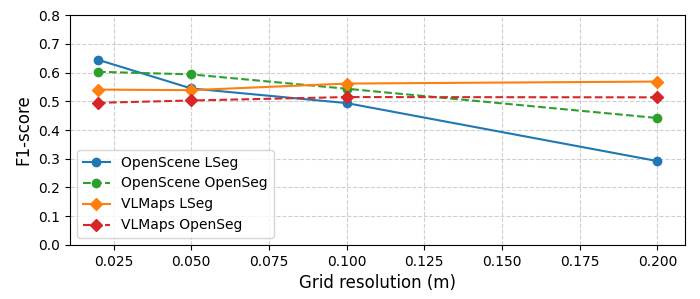}
    \caption{The F1-score of the methods in semantic segmentation task as a function of the map cell size.}
    \label{fig:resolution-segmentation}
    \figvspace{}
\end{figure}

\subsubsection{VLMaps querying}
In the queryability experiment using VLMaps querying, both VLMaps methods performed substantially better than OpenScene. While the results between semantic segmentation and VLMaps querying are not directly comparable, overall low F1-scores indicate that comparison with a single label, \qt{other}, is not sufficient. It is unlikely that any one label would suffice as a universal complement for arbitrary open-vocabulary queries to threshold the results accurately. In a semantic segmentation setting, when the label set is suitable to the data set, the larger set of comparison labels could alleviate this problem by allowing more points of comparison to threshold the query results.

\subsubsection{Prompt engineering}
One approach to improve the matching of text queries proposed in OpenScene and VLMaps is to use prompt engineering. In our experiment, prompt engineering improves the precision of VLMaps methods, yielding an average $0.059$ and $0.008$ increase in the F1-score on LSeg and OpenSeg, respectively. While OpenScene proposes to use a single prompt augmentation \qt{a XX in a scene}, the set of prompt engineering phrases used in VLMaps did not work with OpenScene. This indicates that the encoders are sensitive to prompt engineering and that a single set of phrases might not work with all encoders. Therefore, while prompt engineering could improve the performance, it does not solve the problem of open-vocabulary query thresholding.

\subsubsection{Resolution}
The effect of map voxel size, \ie{} the resolution, to the F1-score is shown in Figure \ref{fig:resolution-segmentation}. We present only the F1-score for presentation clarity, as the precision/recall ratio remains similar over the resolutions. The effect of resolution change is minimal in VLMaps as image features are naturally more scale-invariant, as objects of the same category are perceived from different distances in the training data. The 3D features used by OpenScene are not scale invariant, as the point cloud measurements are on an absolute scale. This leads the performance to decrease rapidly. As the difference between the training resolution and the map resolution grows, the number of out-of-distribution samples grows. This effect is especially pronounced in OpenScene using LSeg.

\subsubsection{Open-vocabulary}
Additionally, we qualitatively demonstrate our evaluation metrics with open-vocabulary queries. The queries, which are only used in this qualitative demonstration, are: \qt{sofa}, \qt{couch}, \qt{white sofa}, \qt{striped}, and \qt{relax}. We use the VLMaps' binary masking method without prompt engineering or dilation as the comparison method. From the results presented in Figure \ref{fig:header-img}, where the matches are shown in orange and non-matches in blue, it can be seen that OpenScene query results are considerably less noisy and encompass entire objects, indicating that the features are more distinctive. While LSeg yields the objects more accurately, the more abstract queries, such as \qt{striped} or \qt{relax}, produce no results. OpenSeg seems to be more sensitive, finding solutions to abstract queries but yielding many false positive results. While this is in line with the findings of the other experimental results, it might hint towards OpenSeg having a broader vocabulary.

\begin{figure}[t]
    \centering
    \includegraphics[width=0.49\textwidth]{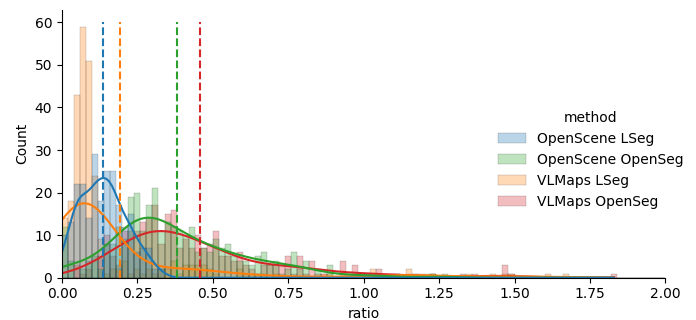}
    \caption{The intra-map distinctness ratios of the methods as histograms, with a kernel density estimation, which is smoothed with a Gaussian kernel. The mean of the distribution is depicted with a vertical line of corresponding color.}
    \label{fig:intra-map-results}
    \figvspace{}
\end{figure}
\newcommand\histImage{0.5}
\newcommand\histDiv{0.49}

\begin{figure*}[ht]
    \begin{subfigure}[t]{\histDiv\textwidth}
        \centering
        \scalebox{\histImage}{\includegraphics{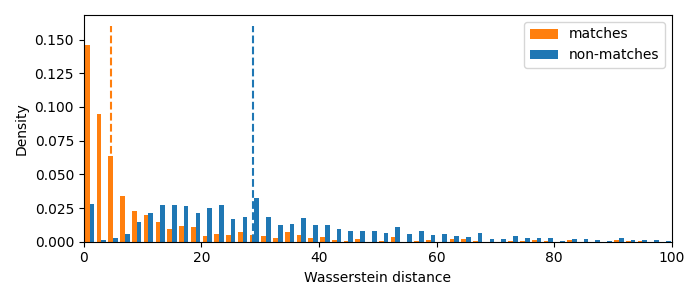}}
        \caption{OpenScene with LSeg}
        \label{fig:inter-openscene-lseg-hist}
    \end{subfigure}
    \begin{subfigure}[t]{\histDiv\textwidth}
        \centering
        \scalebox{\histImage}{\includegraphics{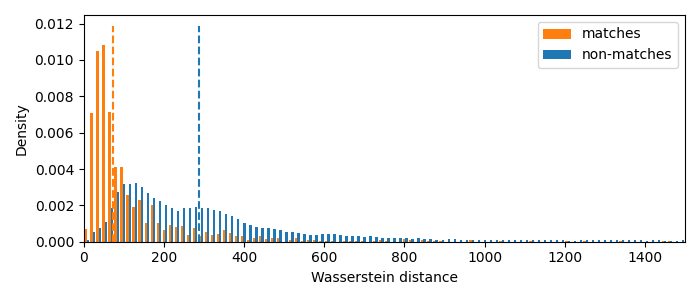}}
        \caption{OpenScene with OpenSeg}
        \label{fig:inter-openscene-openseg-hist}
    \end{subfigure}
    \begin{subfigure}[t]{\histDiv\textwidth}
        \centering
        \scalebox{\histImage}{\includegraphics{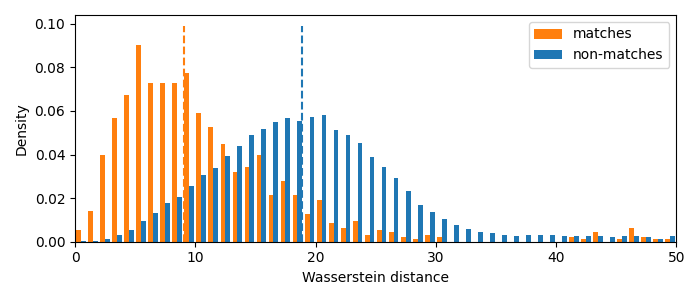}}
        \caption{VLMaps with LSeg}
        \label{fig:inter-vlmaps-lseg-hist}
    \end{subfigure}
    \begin{subfigure}[t]{\histDiv\textwidth}
        \centering
        \scalebox{\histImage}{\includegraphics{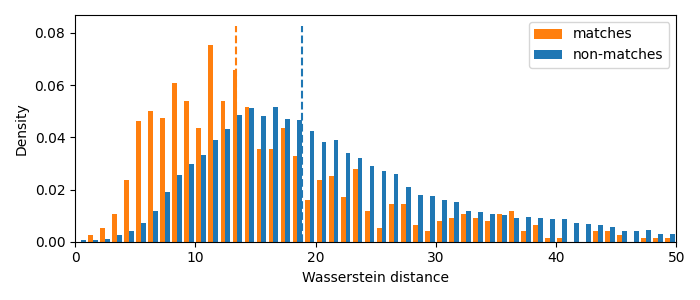}}
        \caption{VLMaps with OpenSeg}
        \label{fig:inter-vlmaps-openseg-hist}
    \end{subfigure}
    
    \caption{The distribution of Wasserstein distances of matching and non-matching labels are shown in orange and blue, respectively. The median of each distribution is presented with a dashed line. }
    \label{fig:inter-map-histogram}
    \figvspace{}
\end{figure*}
\newcommand\interImage{0.5}
\newcommand\interDiv{0.49}

\begin{figure*}[ht]
    \begin{subfigure}[t]{\interDiv\textwidth}
        \centering
        \scalebox{\interImage}{\includegraphics{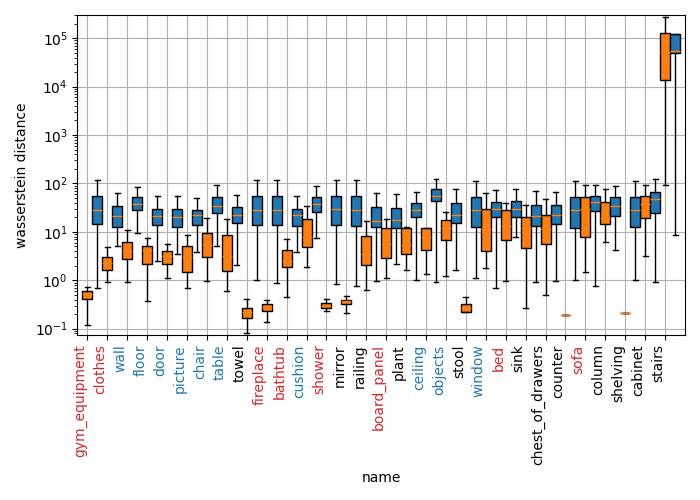}}
        \caption{OpenScene LSeg}
        \label{fig:inter-openscene-lseg}
    \end{subfigure}
    \begin{subfigure}[t]{\interDiv\textwidth}
        \centering
        \scalebox{\interImage}{\includegraphics{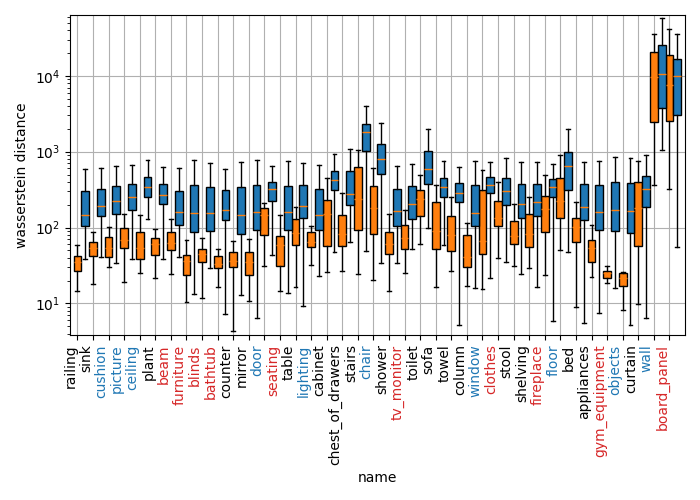}}
        \caption{OpenScene OpenSeg}
        \label{fig:inter-openscene-openseg}
    \end{subfigure}
    \begin{subfigure}[t]{\interDiv\textwidth}
        \centering
        \scalebox{\interImage}{\includegraphics{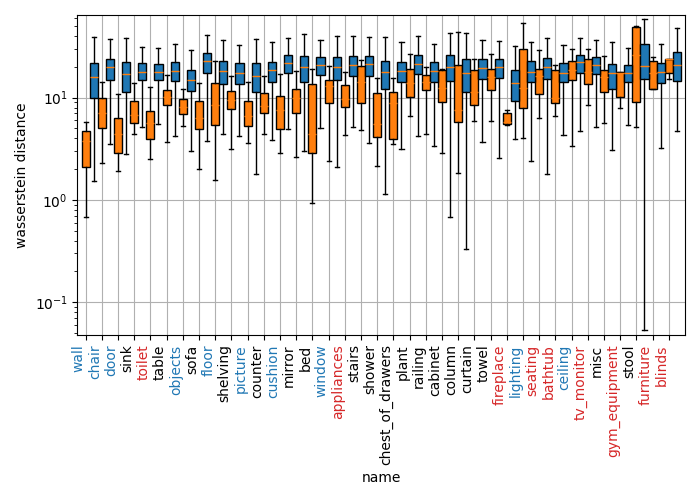}}
        \caption{VLMaps LSeg}
        \label{fig:inter-vlmaps-lseg}
    \end{subfigure}
    \begin{subfigure}[t]{\interDiv\textwidth}
        \centering
        \scalebox{\interImage}{\includegraphics{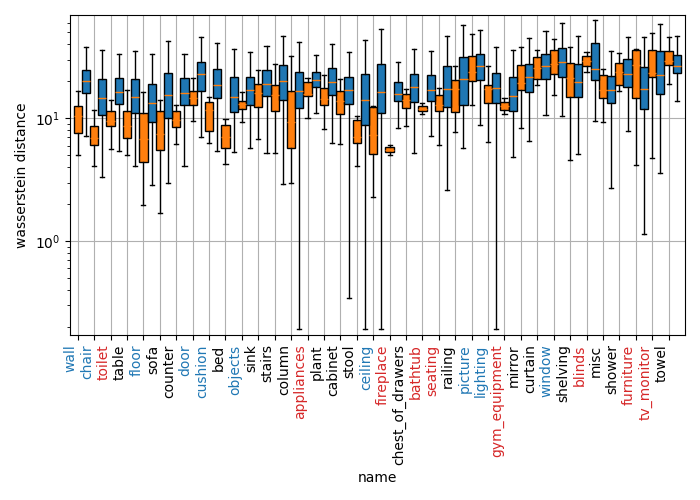}}
        \caption{VLMaps OpenSeg}
        \label{fig:inter-vlmaps-openseg}
    \end{subfigure}
    
    \caption{The Wasserstein distances of the matching labels are depicted as orange bars and non-matching labels as blue bars on a logarithmic scale. The ten most common and uncommon labels are marked with red and blue text, respectively. The labels are sorted per image on the separability of the distance distribution according to Krusal-Wallis statistic \cite{kruskal1952use}. The outliers of the boxplots have been omitted for clarity. The OpenScene distances, in (a) and (b), are multiple orders of magnitude larger, as the embeddings are an order of magnitude larger.}
    \label{fig:inter-map-results}
    \figvspace{}
\end{figure*}

\subsection{Distinctness}
\label{sec:distinctness}

\subsubsection{Intra-map distinctness} 
\label{sec:intra-map} 
The results are presented in Figure \ref{fig:intra-map-results}, in which the intra-map distinctness ratios of the methods are illustrated with histograms. Additionally, the means of data are shown with dashed lines. LSeg has better distinctness between the labels with each method, which is indicated by the lower means of the distributions. This indicates that LSeg features of semantic classes are more distinct from each other than in OpenSeg. Furthermore, the 3D ensemble features of OpenScene help to make the features slightly more distinct, as OpenScene has lower means than VLMaps with comparable encoders. 

More distinct features, with lower intra-map distinctness ratios, should be preferable, as that indicates better queryability performance. The intuition is that in a best-case scenario, where the query is very accurate, \ie{} close to the desired map features in the embedding space, tightly clustered classes are retrieved as a whole, which should lead to better results. More spread-out embeddings are more likely to return partial and sporadic results. 

There are multiple classes with a ratio above one, which means that the embeddings in those classes are more different from each other than the map on average. This implies that these classes are challenging to cluster and hard to distinguish. While OpenScene with LSeg does not have any class with a ratio over $1$, OpenScene with OpenSeg has five: \qt{fireplace}, \qt{lighting}, \qt{beam}, \qt{lighting}, \qt{shower}. These classes amount to, on average, $0.8\%$ number of voxels in the whole data set and are, therefore, rare tail classes. VLMaps with both encoders have a larger number of classes with a ratio over one. Notably, both include large classes \qt{floor} and \qt{table}, consisting of $4.44\%$ and $1.67\%$ of the data set, respectively. This implies that spatial features could be helpful in distinguishing large background classes, such as \qt{floor}.

\subsubsection{Inter-map distinctness} 
\label{sec:inter-map} 
In Figure \ref{fig:inter-map-histogram}, the Wasserstein distances between matching, \ie{} the label is the same, and non-matching sets of embeddings across all maps are shown, \ie{} the blue distances show the similarity between embeddings of the same label between different maps, whereas the orange distances are similarity between embeddings of different classes. The medians of the data are shown with dashed lines. Again, a larger separation between matching and non-matching is preferable, as this would imply that classes remain distinct between environments, \ie{} a door would look more like other doors than other classes.

The ratio between the medians of matching and non-matching labels suggests that OpenScene separates the labels better than VLMaps, and LSeg separates them better than OpenSeg. The ratio of medians of OpenScene with LSeg is 6.31, OpenScene with OpenSeg is 3.78, VLMaps with LSeg 2.07, and VLMaps with OpenSeg 1.42, which is the same ordering observed in queryability results in Section \ref{sec:queryability}.

OpenScene has larger Wasserstein distances on average, even with matching labels, which results from the fact that the 3D features of OpenSeg are of a larger magnitude than the image features of both VLMaps and OpenScene image features. For example, the average norm of all OpenScene with OpenSeg embeddings in the map of the first sequence is 13.68 times larger than the average norm of all VLMaps with OpenSeg embeddings in the same map, and the maximum being 57.97 times larger than the maximum VLMaps feature.

Furthermore, in Figure \ref{fig:inter-map-results}, matching and non-matching distances are shown per class with orange and blue bars, respectively. The same phenomena can be seen: LSeg distinguishes the classes better than OpenSeg and OpenScene better than VLMaps, which is indicated by the separation of the blue and orange box plots. Additionally, the ten most common and uncommon categories in the Matterport3D data set that are present in each set of predictions are marked with blue and red names, respectively. VLMaps performs well in separating common classes and worse with uncommon classes, especially using LSeg, whereas OpenScene performs better across the whole set of labels.

When considering all these results, the combination of OpenScene and LSeg produces maps that are more queryable and distinct, making it the method that can currently be expected to lead to more accurate performance in downstream robotic applications. However, OpenScene must be used with the resolution that the 3D encoder is trained on, whereas VLMaps can generalize better to different robot hardware, as it can produce maps of different resolutions, allowing the creation of maps with small memory footprint, with no significant reduction in quality.

\section{Conclusion}
\label{sec:conclusion}

In this work, we evaluate the quality of \sota{} latent semantic maps. We argue that success in downstream tasks is not alone sufficient evidence of the quality of the maps, but rather the representations themselves should be evaluated to understand their limitations. Therefore, we propose quantitative metrics to enable a fair comparison between methods. 

In our experiments,  OpenScene mapping using 3D scene structure with LSeg semantic embedding performed the best, which indicates that the 3D structure of the environment should be used for inferring the latent semantics. While LSeg performed best in the semantic segmentation setting, the qualitative demonstration hints that OpenSeg might be more sensitive to rarer and more abstract queries when properly thresholded. Furthermore, while 2D image features are naturally scale-invariant, the 3D features are susceptible to resolution changes. In use cases with limited hardware resources, the image-based methods work with various resolutions, whereas 3D features must be trained for each resolution. This improves the adaptability of image-based methods to various use cases in real-world settings.

The ability of visual-language maps to capture latent semantics is, however, still limited. Currently, fully supervised semantic segmentation methods outperform all latent semantic maps. The open-vocabulary querying method, proposed in VLMaps, comparing the query to the string \qt{other} cannot sufficiently partition the environment, yielding poor results.  This means that a practitioner must choose between the accuracy of the fine-tuned methods with more limited semantics and the broad vocabulary of the less accurate open-vocabulary methods. The relatively low F1-scores and \glspl{iou} indicate that in open-vocabulary settings, the thresholding and clustering of the query results remain an open problem.

The growing adoption of latent semantic maps in robotics is already allowing more natural human-robot interaction and complex mobile task execution. The results presented in this work provide evidence that rigorously studying the qualities of these representations is important both to inform practitioners integrating these maps in their robotic pipelines and outlining open problems for researchers to explore.

\bibliographystyle{IEEEtran}
\bibliography{clean-abbr,IEEEabrv,ctrl}

\begin{thebibliography}{10}
\providecommand{\url}[1]{#1}
\csname url@samestyle\endcsname
\providecommand{\newblock}{\relax}
\providecommand{\bibinfo}[2]{#2}
\providecommand{\BIBentrySTDinterwordspacing}{\spaceskip=0pt\relax}
\providecommand{\BIBentryALTinterwordstretchfactor}{4}
\providecommand{\BIBentryALTinterwordspacing}{\spaceskip=\fontdimen2\font plus
\BIBentryALTinterwordstretchfactor\fontdimen3\font minus
  \fontdimen4\font\relax}
\providecommand{\BIBforeignlanguage}[2]{{%
\expandafter\ifx\csname l@#1\endcsname\relax
\typeout{** WARNING: IEEEtran.bst: No hyphenation pattern has been}%
\typeout{** loaded for the language `#1'. Using the pattern for}%
\typeout{** the default language instead.}%
\else
\language=\csname l@#1\endcsname
\fi
#2}}
\providecommand{\BIBdecl}{\relax}
\BIBdecl

\bibitem{tenorth_knowrob_2009}
M.~Tenorth and M.~Beetz, ``\BIBforeignlanguage{en}{{KNOWROB} - knowledge
  processing for autonomous personal robots},'' in
  \emph{\BIBforeignlanguage{en}{Proc. {IEEE}/{RSJ} {Int.} {Conf.} on {Intell.}
  {Robots} and {Syst.} (IROS)}}, St. Louis, MO, USA, Oct. 2009, pp. 4261--4266.

\bibitem{bendale_towards_2016}
A.~Bendale and T.~E. Boult, ``\BIBforeignlanguage{en}{Towards {Open} {Set}
  {Deep} {Networks}},'' in \emph{\BIBforeignlanguage{en}{Proc. {IEEE} {Conf.}
  on {Comput.} {Vis.} and {Pattern} {Recognit.} {(CVPR)}}}, Las Vegas, NV, USA,
  Jun. 2016, pp. 1563--1572.

\bibitem{geng_recent_2021}
C.~Geng, S.-J. Huang, and S.~Chen, ``\BIBforeignlanguage{en}{Recent {Advances}
  in {Open} {Set} {Recognition}: {A} {Survey}},''
  \emph{\BIBforeignlanguage{en}{IEEE Trans. on Pattern Anal. and Mach.
  Intell.}}, vol.~43, no.~10, pp. 3614--3631, Oct. 2021.

\bibitem{huang_visual_2023}
C.~Huang, O.~Mees, A.~Zeng, and W.~Burgard, ``\BIBforeignlanguage{en}{Visual
  {Language} {Maps} for {Robot} {Navigation}},'' in
  \emph{\BIBforeignlanguage{en}{Proc. {IEEE} {Int.} {Conf.} on {Robot.} and
  {Automat.} {(ICRA)}}}, London, UK, May 2023, pp. 10\,608--10\,615.

\bibitem{gadre_cows_2023}
S.~Y. Gadre, M.~Wortsman, G.~Ilharco, L.~Schmidt, and S.~Song, ``Cows on
  pasture: Baselines and benchmarks for language-driven zero-shot object
  navigation,'' in \emph{Proc. IEEE/CVF Conf. on Comput. Vis. and Pattern
  Recognit. (CVPR)}, Vancouver, Canada, Jun. 2023, pp. 23\,171--23\,181.

\bibitem{shah_lm-nav_2022}
D.~Shah, B.~Osi{\'n}ski, B.~Ichter, and S.~Levine, ``Lm-nav: Robotic navigation
  with large pre-trained models of language, vision, and action,'' in
  \emph{Proc. Conf. on Robot Learn. (CoRL)}, Atlanta, GA, USA, Nov. 2023, pp.
  492--504.

\bibitem{peng_openscene_2023}
S.~Peng \emph{et~al.}, ``Openscene: 3d scene understanding with open
  vocabularies,'' in \emph{Proc. of the IEEE/CVF Conf. on Comput. Vis. and
  Pattern Recognit. (CVPR)}, Vancouver, Canada, Jun. 2023, pp. 815--824.

\bibitem{radford_learning_2021}
A.~Radford \emph{et~al.}, ``Learning transferable visual models from natural
  language supervision,'' in \emph{Proc. Int. Conf. on Mach. Learn. (ICML)},
  Virtual, Jul. 2021, pp. 8748--8763.

\bibitem{chen_open-vocabulary_2022}
B.~Chen \emph{et~al.}, ``Open-vocabulary queryable scene representations for
  real world planning,'' in \emph{Proc. IEEE Int. Conf. on Robot. and Automat.
  (ICRA)}, London, UK, May 2023, pp. 11\,509--11\,522.

\bibitem{yuan_uni-fusion_2024}
Y.~Yuan and A.~Nüchter, ``Uni-fusion: Universal continuous mapping,''
  \emph{IEEE Trans. on Robot. (T-RO)}, vol.~40, pp. 1373--1392, Jan. 2024.

\bibitem{jatavallabhula23conceptfusion}
K.~Jatavallabhula \emph{et~al.}, ``Conceptfusion: Open-set multimodal 3d
  mapping,'' in \emph{Proc. Robotics: Sci. and Syst. (RSS)}, Daegu, Korea, Jul.
  2023.

\bibitem{gu2024conceptgraphs}
Q.~Gu \emph{et~al.}, ``\BIBforeignlanguage{en}{Conceptgraphs: Open-vocabulary
  3d scene graphs for perception and planning},'' in
  \emph{\BIBforeignlanguage{en}{Proc. {IEEE} {Int.} {Conf.} on {Robot.} and
  {Automat.} {(ICRA)}}}, Yokohama, Japan, May 2024, pp. 5021--5028.

\bibitem{werby24hovsg}
A.~Werby, C.~Huang, M.~Büchner, A.~Valada, and W.~Burgard, ``Hierarchical
  open-vocabulary 3d scene graphs for language-grounded robot navigation,'' in
  \emph{Proc. Robotics: Sci. and Syst. (RSS)}, Delft, The Netherlands, Jul.
  2024.

\bibitem{li_language-driven_2022}
B.~Li, K.~Q. Weinberger, S.~Belongie, V.~Koltun, and R.~Ranftl,
  ``\BIBforeignlanguage{en}{Language-driven {Semantic} {Segmentation}},'' Apr.
  2022, arXiv:2201.03546 [cs].

\bibitem{ghiasi_scaling_2022}
G.~Ghiasi, X.~Gu, Y.~Cui, and T.-Y. Lin, ``Scaling open-vocabulary image
  segmentation with image-level labels,'' in \emph{Proc. Eur. Conf. on Comput.
  Vis. (ECCV)}, Tel Aviv, Israel, Oct. 2022, pp. 540--557.

\bibitem{zhao_open_2017}
H.~Zhao, X.~Puig, B.~Zhou, S.~Fidler, and A.~Torralba,
  ``\BIBforeignlanguage{en}{Open {Vocabulary} {Scene} {Parsing}},'' in
  \emph{\BIBforeignlanguage{en}{Proc. {IEEE} {Int.} {Conf.} on {Comput.} {Vis.}
  {(ICCV)}}}, Venice, Oct. 2017, pp. 2021--2029.

\bibitem{bravo_open-vocabulary_2023}
M.~A. Bravo, S.~Mittal, S.~Ging, and T.~Brox,
  ``\BIBforeignlanguage{en}{Open-vocabulary {Attribute} {Detection}},'' in
  \emph{\BIBforeignlanguage{en}{Proc. {IEEE}/{CVF} {Conf.} on {Comput.} {Vis.}
  and {Pattern} {Recognit.} {(CVPR)}}}, Vancouver, BC, Canada, Jun. 2023, pp.
  7041--7050.

\bibitem{bansal_zero-shot_2018}
A.~Bansal, K.~Sikka, G.~Sharma, R.~Chellappa, and A.~Divakaran,
  ``\BIBforeignlanguage{en}{Zero-{Shot} {Object} {Detection}},'' in
  \emph{\BIBforeignlanguage{en}{Proc. Comput. {Vis.} {ECCV}}}, vol. 11205,
  Cham, Sep. 2018, pp. 397--414.

\bibitem{zareian_open-vocabulary_2021}
A.~Zareian, K.~D. Rosa, D.~H. Hu, and S.-F. Chang,
  ``\BIBforeignlanguage{en}{Open-{Vocabulary} {Object} {Detection} {Using}
  {Captions}},'' in \emph{\BIBforeignlanguage{en}{Proc. {IEEE}/{CVF} {Conf.} on
  {Comput.} {Vis.} and {Pattern} {Recognit.} {(CVPR)}}}, Nashville, TN, USA,
  Jun. 2021, pp. 14\,388--14\,397.

\bibitem{gu_open-vocabulary_2022}
X.~Gu, T.-Y. Lin, W.~Kuo, and Y.~Cui, ``\BIBforeignlanguage{en}{Open-vocabulary
  {Object} {Detection} via {Vision} and {Language} {Knowledge}
  {Distillation}},'' May 2022, arXiv:2104.13921 [cs].

\bibitem{feng_promptdet_2022}
C.~Feng \emph{et~al.}, ``Promptdet: Towards open-vocabulary detection using
  uncurated images,'' in \emph{Proc. Eur. Conf. on Comput. Vis. (ECCV)}, Tel
  Aviv, Israel, Oct. 2022, pp. 701--717.

\bibitem{wagan_map_2008}
A.~I. Wagan, A.~Godil, and X.~Li, ``\BIBforeignlanguage{en}{Map quality
  assessment},'' in \emph{\BIBforeignlanguage{en}{Proc. of the 8th {Workshop}
  on {Perform.} {Metrics} for {Intell.} {Syst.}}}, Gaithersburg, MD, USA, Aug.
  2008, pp. 278--282.

\bibitem{kucner_robust_2021}
T.~P. Kucner, M.~Luperto, S.~Lowry, M.~Magnusson, and A.~J. Lilienthal,
  ``\BIBforeignlanguage{en}{Robust {Frequency}-{Based} {Structure}
  {Extraction}},'' in \emph{\BIBforeignlanguage{en}{Proc. {IEEE} {Int.} {Conf.}
  on {Robot.} and {Automat.} {(ICRA)}}}, Xi'an, China, May 2021, pp.
  1715--1721.

\bibitem{aravecchia_comparing_2024}
S.~Aravecchia, M.~Clausel, and C.~Pradalier,
  ``\BIBforeignlanguage{en}{Comparing metrics for evaluating {3D} map quality
  in natural environments},'' \emph{\BIBforeignlanguage{en}{Robot. and Auton.
  Syst. (RAS)}}, vol. 173, p. 104617, Mar. 2024.

\bibitem{salas-moreno_slam_2013}
R.~F. Salas-Moreno, R.~A. Newcombe, H.~Strasdat, P.~H. Kelly, and A.~J.
  Davison, ``\BIBforeignlanguage{en}{{SLAM}++: {Simultaneous} {Localisation}
  and {Mapping} at the {Level} of {Objects}},'' in
  \emph{\BIBforeignlanguage{en}{Proc. {IEEE} {Conf.} on {Comput.} {Vis.} and
  {Pattern} {Recognit.} (CVPR)}}, Portland, OR, USA, Jun. 2013, pp. 1352--1359.

\bibitem{mccormac_fusion_2018}
J.~Mccormac, R.~Clark, M.~Bloesch, A.~Davison, and S.~Leutenegger,
  ``\BIBforeignlanguage{en}{Fusion++: {Volumetric} {Object}-{Level} {SLAM}},''
  in \emph{\BIBforeignlanguage{en}{Proc. {Int.} {Conf.} on {3D} {Vis.}
  {(3DV)}}}, Verona, Italy, Sep. 2018, pp. 32--41.

\bibitem{runz_maskfusion_2018}
M.~Runz, M.~Buffier, and L.~Agapito, ``\BIBforeignlanguage{en}{{MaskFusion}:
  {Real}-{Time} {Recognition}, {Tracking} and {Reconstruction} of {Multiple}
  {Moving} {Objects}},'' in \emph{\BIBforeignlanguage{en}{Proc. {IEEE} {Int.}
  {Symp.} on {Mixed} and {Augmented} {Reality} {(ISMAR)}}}, Munich, Germany,
  Oct. 2018, pp. 10--20.

\bibitem{sunderhauf_meaningful_2017}
N.~Sunderhauf, T.~T. Pham, Y.~Latif, M.~Milford, and I.~Reid,
  ``\BIBforeignlanguage{en}{Meaningful maps with object-oriented semantic
  mapping},'' in \emph{\BIBforeignlanguage{en}{Proc. {IEEE}/{RSJ} {Int.}
  {Conf.} on {Intell.} {Robots} and {Syst.} {(IROS)}}}, Vancouver, BC, Sep.
  2017, pp. 5079--5085.

\bibitem{chen_leveraging_2023}
W.~Chen, S.~Hu, R.~Talak, and L.~Carlone, ``\BIBforeignlanguage{en}{Leveraging
  {Large} ({Visual}) {Language} {Models} for {Robot} {3D} {Scene}
  {Understanding}},'' Nov. 2023, arXiv:2209.05629 [cs].

\bibitem{narita_panopticfusion_2019}
G.~Narita, T.~Seno, T.~Ishikawa, and Y.~Kaji,
  ``\BIBforeignlanguage{en}{{PanopticFusion}: {Online} {Volumetric} {Semantic}
  {Mapping} at the {Level} of {Stuff} and {Things}},'' in
  \emph{\BIBforeignlanguage{en}{Proc. {IEEE}/{RSJ} {Int.} {Conf.} on {Intell.}
  {Robots} and {Syst.} {(IROS)}}}, Macau, China, Nov. 2019, pp. 4205--4212.

\bibitem{chang_matterport3d_2017}
A.~Chang \emph{et~al.}, ``\BIBforeignlanguage{en}{{Matterport3D}: {Learning}
  from {RGB}-{D} {Data} in {Indoor} {Environments}},'' Sep. 2017,
  arXiv:1709.06158 [cs].

\bibitem{adams_seeded_1994}
R.~Adams and L.~Bischof, ``\BIBforeignlanguage{en}{Seeded region growing},''
  \emph{\BIBforeignlanguage{en}{IEEE Trans. on Pattern Anal. and Mach.
  Intell.}}, vol.~16, no.~6, pp. 641--647, Jun. 1994.

\bibitem{VLMaps_software}
\BIBentryALTinterwordspacing
C.~Huang, O.~Mees, A.~Zeng, and W.~Burgard. Vlmaps. [Online]. Available:
  \url{{https://github.com/vlmaps/vlmaps}}
\BIBentrySTDinterwordspacing

\bibitem{habitat19iccv}
M.~Savva \emph{et~al.}, ``Habitat: A platform for embodied ai research,'' in
  \emph{Proc. IEEE/CVF Int. Conf. on Comput. Vis. (ICCV)}, Seoul, Korea, Oct.
  2019, pp. 9338--9346.

\bibitem{szot2021habitat}
A.~Szot \emph{et~al.}, ``Habitat 2.0: Training home assistants to rearrange
  their habitat,'' in \emph{Advances in Neural Inf. Process. Syst. (NeurIPS)},
  vol.~34, Virtual, Dec. 2021, pp. 251--266.

\bibitem{puig2023habitat3}
X.~Puig \emph{et~al.}, ``Habitat 3.0: A co-habitat for humans, avatars and
  robots,'' 2023, arXiv:2310.13724 [cs].

\bibitem{kruskal1952use}
W.~H. Kruskal and W.~A. Wallis, ``Use of ranks in one-criterion variance
  analysis,'' \emph{J. of the Amer. statistical Assoc.}, vol.~47, no. 260, pp.
  583--621, Apr. 1952.

\end{thebibliography}
\end{document}